\documentclass[10pt,twocolumn,letterpaper]{article}

\usepackage{cvpr}              % To produce the CAMERA-READY version
\definecolor{cvprblue}{rgb}{0.21,0.49,0.74}
\usepackage[pagebackref,breaklinks,colorlinks,allcolors=cvprblue]{hyperref}

\usepackage{multirow}
\usepackage[table]{xcolor}
\usepackage{amssymb}

\def\paperID{*****} % *** Enter the Paper ID here
\def\confName{CVPR}
\def\confYear{2025}

\title{Dual-Stage Reweighted MoE for Long-Tailed Egocentric Mistake Detection}

\author{
	Boyu Han\textsuperscript{1,2}\hspace{2em} Qianqian Xu\textsuperscript{1,}\thanks{Corresponding authors}\hspace{2em} Shilong Bao\textsuperscript{2}\hspace{2em} Zhiyong Yang\textsuperscript{2}\hspace{2em} \\Sicong Li\textsuperscript{3,4}\hspace{2em} Qingming Huang\textsuperscript{2,1,*} \\
	{\textsuperscript{1}Key Lab. of Intelligent Information Processing, Institute of Computing Technology, CAS} \\
	{\textsuperscript{2}School of Computer Science and Tech., University of Chinese Academy of Sciences} \\
    {\textsuperscript{3}Institute of Information Engineering, CAS} \\
    {\textsuperscript{4}School of Cyber Security, University of Chinese Academy of Sciences} \\
	{\tt\small \{hanboyu23z, xuqianqian\}@ict.ac.cn\hspace{2em} lisicong24@mails.ucas.ac.cn}\\ 
    {\tt\small \{baoshilong, yangzhiyong21, qmhuang\}@ucas.ac.cn}
}

\begin{document}
\maketitle
\begin{abstract}
In this report, we address the problem of determining whether a user performs an action incorrectly from egocentric video data. To handle the challenges posed by subtle and infrequent mistakes, we propose a Dual-Stage Reweighted Mixture-of-Experts (DR-MoE) framework. In the first stage, features are extracted using a frozen ViViT model and a LoRA-tuned ViViT model, which are combined through a feature-level expert module. In the second stage, three classifiers are trained with different objectives: reweighted cross-entropy to mitigate class imbalance, AUC loss to improve ranking under skewed distributions, and label-aware loss with sharpness-aware minimization to enhance calibration and generalization. Their predictions are fused using a classification-level expert module. The proposed method achieves strong performance, particularly in identifying rare and ambiguous mistake instances. The code is available at \href{https://github.com/boyuh/DR-MoE}{https://github.com/boyuh/DR-MoE}.
\end{abstract}    
\section{Introduction}
\label{sec:introduction}

This document introduces the solution proposed by the MR-CAS team for the Mistake Detection Challenge of the HoloAssist 2025 competition~\cite{wang2023holoassist}. The objective of the task is to determine whether a user makes a mistake while performing an operation, such as assembling a switch. Unlike traditional action recognition and model-level error detection~\cite{liang2024badclip,zhang2024towards,lin2025ltd,liu2024multimodal}, mistake detection in this work focuses on assessing the quality of the action execution. This requires a more detailed temporal analysis and heightened sensitivity to subtle deviations in user behavior.

In practical terms, this problem is challenging due to two main factors. First, mistake events are inherently rare and often ambiguous, leading to severely imbalanced data distributions that hinder conventional supervised learning. Second, the diversity of task types and user behaviors in real-world egocentric videos introduces significant intra-class variation, making it difficult for a single model to generalize effectively~\cite{liu2024not}.

To address these challenges, we propose a \textit{Dual-Stage Reweighted Mixture-of-Experts (DR-MoE)} framework designed to exploit complementary modeling strategies at both the feature extraction and classification levels. In the first stage, we extract spatiotemporal representations using two ViViT-based~\cite{arnab2021vivit} experts: one frozen to preserve generic semantic priors and one fine-tuned with low-rank adaptation (LoRA)~\cite{hulora} to focus on mistake-sensitive cues. These representations are fused through a learnable Feature Mixture-of-Experts (F-MoE) module, allowing dynamic weighting based on input characteristics.

\begin{figure*}
  \centering
  \includegraphics[width=\linewidth]{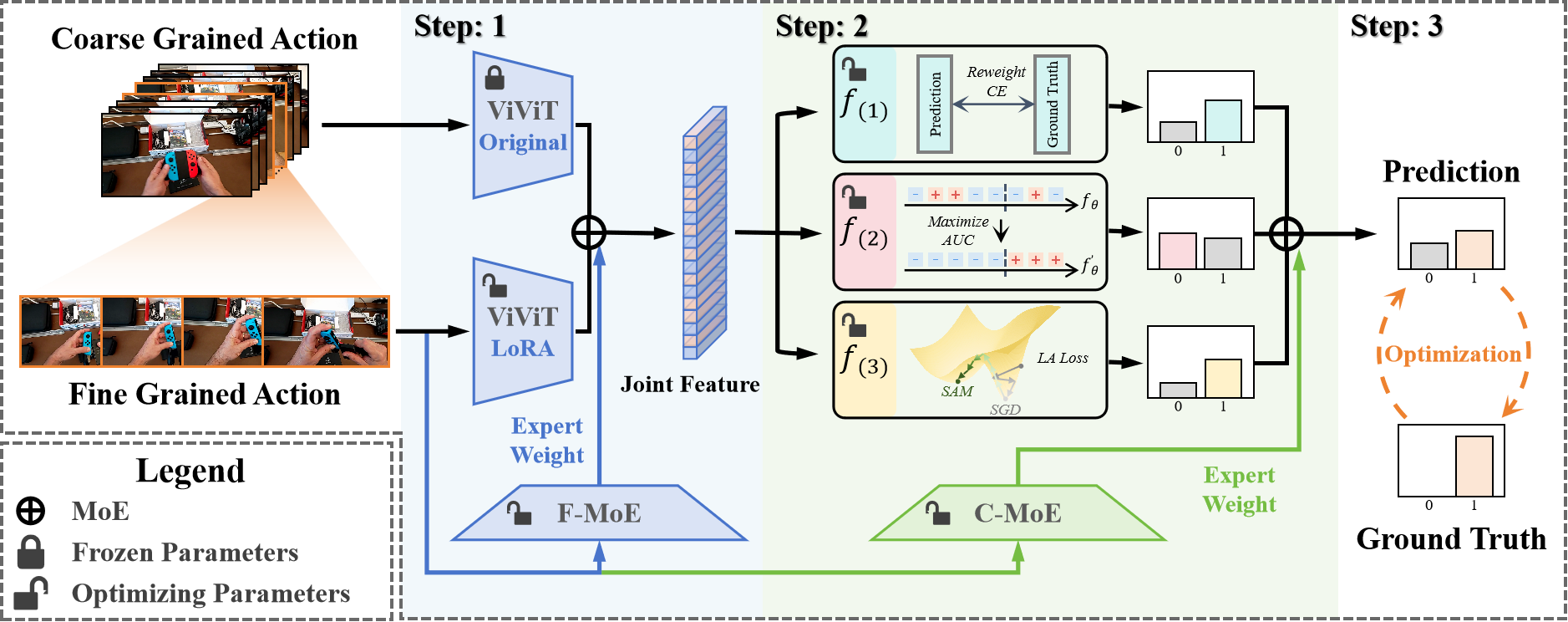}
  \caption{\textbf{An overview of our DR-MoE method.} Our method consists of two main stages. The first stage performs feature extraction by combining a frozen ViViT backbone and a LoRA-tuned ViViT backbone through a Feature Mixture-of-Experts (F-MoE) module. The second stage carries out classification by integrating the outputs of three independently trained classifiers using a Classification Mixture-of-Experts (C-MoE) module to generate the final prediction.}
  \label{fig: overview}
\end{figure*}

In the second stage, the fused representation is passed through three independently optimized classifiers, each trained with a different objective tailored to long-tailed recognition: class-rebalanced cross-entropy~\cite{cui2019class}, AUC maximization~\cite{yang2021learning}, and label-aware (LA) loss~\cite{menon2020long} enhanced with sharpness-aware minimization (SAM)~\cite{foretsharpness,li2025focal}. The reweighted cross-entropy loss addresses severe class imbalance by assigning higher importance to rare mistake classes. The AUC-based objective focuses on improving ranking quality between positive and negative instances, which is crucial when the error class is underrepresented~\cite{hanley1982meaning,hanaucseg,bao2024improved,huaopenworldauc,lisize}. The LA loss with SAM promotes more calibrated and robust decision boundaries by adjusting logits and explicitly optimizing for flat minima. The outputs of these classifiers are adaptively integrated using a Classification Mixture-of-Experts (C-MoE) module, which dynamically weighs each expert’s contribution based on the input, enabling flexible decision-making under diverse data conditions.

By leveraging diverse inductive biases from both backbone models and loss formulations, our approach significantly improves performance on mistake recognition, especially in underrepresented scenarios. The following sections detail our method, its implementation, and evaluation on the HoloAssist benchmark.

\section{Method}
\label{sec:method}

\subsection{Overview}
\label{subsec:Overview}
The core objective of mistake detection is to determine whether a user makes an error while performing a specific activity in a video. Specifically, we are given a coarse-grained action video $V_{0:T}$, where $T$ denotes the total duration of the video. A fine-grained action segment $V_{t:t+\tau}$ represents the user executing a particular task, with $t \geq 0$, $t+\tau \leq T$, and $T > 0$. The model is responsible for determining whether the fine-grained action contains a mistake, outputting $1$ for a mistake and $0$ for a correct action.

\Cref{fig: overview} provides an overview of our approach. It consists of two key stages. \textbf{In the first stage}, we extract features using both a frozen ViViT~\cite{arnab2021vivit} backbone and a ViViT model fine-tuned with LoRA~\cite{hulora}. These two feature extractors act as experts, and their outputs are combined through a \textit{Feature Mixture-of-Experts (F-MoE)} module to generate a unified joint representation. \textbf{In the second stage}, the joint features are fed into three separate classification heads, each optimized with a different loss function: Reweight CE Loss~\cite{cui2019class}, AUC Loss~\cite{yang2021learning}, and LA Loss~\cite{menon2020long} with SAM~\cite{foretsharpness}. The predictions from these heads are then adaptively fused by a \textit{Classification Mixture-of-Experts (C-MoE)} module to produce the final output. Applying Mixture-of-Experts~\cite{yang2024harnessing,li2025hybrid,hua2024reconboost} at both the feature extraction and classification stages allows the model to integrate the complementary strengths of each expert, thereby enhancing robustness and performance, particularly in challenging, long-tailed mistake detection scenarios.

\subsection{Feature Mixture of Experts (F-MoE)}
\label{subsec:F-MoE}
We extract video representations using two complementary ViViT-based backbones. The ViViT model, pretrained on the Kinetics dataset~\cite{kay2017kinetics}, excels at capturing coarse-grained human action features. We directly utilize a frozen ViViT model $f_{\text{ori}}$ to extract these generic spatiotemporal patterns from the input video segment $V_{0:T}$. However, since the mistake detection task requires finer-grained action understanding that is not covered by the pre-training, we apply LoRA to fine-tune the model. The fine-tuned model $f_{\text{lora}}$ is then used to extract features tailored to the specific requirements of mistake detection.

To effectively combine the strengths of both backbones, we introduce a Feature Mixture-of-Experts (F-MoE) module that fuses their outputs through a learnable weighting mechanism. Specifically, the joint feature representation is computed as
\begin{equation}
    \mathbf{F}_{\text{joint}} = \alpha \cdot f_{\text{ori}}(V_{0:T}) + (1 - \alpha) \cdot f_{\text{lora}}(V_{t:t+\tau}),
\end{equation}
where $\alpha \in [0, 1]$ is an adaptive parameter that adjusts the contribution of each expert. This approach allows the network to dynamically prioritize the more informative source based on the characteristics of each input instance, thereby enhancing its generalization ability across both typical and error-prone scenarios.

\subsection{Classification Mixture of Experts (C-MoE)}
\label{subsec:C-MoE}

\begin{table*}[t!]
  \centering
  \renewcommand\arraystretch{1}
  \caption{\textbf{Results obtained on the test set.} The champion and the runner-up are highlighted in \textbf{bold} and \underline{underline}.}
  \resizebox{0.8\linewidth}{!}{
    \begin{tabular}{ccccccc}
    \toprule
    \multirow{2}[2]{*}{\textbf{Method}} & \multirow{2}[2]{*}{\textbf{Modality}} & \multirow{2}[2]{*}{\textbf{F-score}} & \multicolumn{2}{c}{\textbf{Correct}} & \multicolumn{2}{c}{\textbf{Mistake}} \\
          &       &       & \textbf{Precision} & \textbf{Recall} & \textbf{Precision} & \textbf{Recall} \\
    \midrule
    Random & -     & 0.28  & 0.61  & 0.10  & \textbf{0.15} & 0.46  \\
    \midrule
    \multicolumn{1}{c}{\multirow{4}[2]{*}{TimeSformer \textsubscript{\textcolor[rgb]{ .933,  .51,  .184}{(Baseline)}}}} & RGB   & 0.35  & 0.83  & 0.52  & \underline{0.13}  & 0.27  \\
          & Hands & 0.40  & 0.93  & 0.52  & \underline{0.13}  & 0.31  \\
          & RGB+Hands & 0.36  & 0.86  & 0.43  & 0.10  & 0.12  \\
          & RGB+Hands+Eyes & 0.32  & 0.89  & 0.43  & 0.11  & \underline{0.50}  \\
    \midrule
    \multicolumn{1}{c}{UNICT Solution \textsubscript{\textcolor[rgb]{ .933,  .51,  .184}{(2024 Top1)}}} & RGB+Eyes & \underline{0.51}  & \underline{0.95}  & \textbf{0.93} & 0.06  & 0.09  \\
    \midrule
    \rowcolor[rgb]{ .992,  .91,  .855} DR-MoE \textsubscript{\textcolor[rgb]{ .933,  .51,  .184}{(Ours)}}  & RGB   & \textbf{0.57} & \textbf{0.97} & \underline{0.60}  & 0.08  & \textbf{0.63} \\
    \bottomrule
    \end{tabular}%
    }
  \label{tab: results}%
\end{table*}%

Once the joint feature $\mathbf{F}_{\text{joint}}$ is obtained, it is passed through a classification module to determine whether the observed action contains a mistake. However, due to the diversity of video types and the varying degrees of class imbalance inherent in the HoloAssist datasets~\cite{wang2023holoassist}, relying on a single classification head often fails to achieve satisfactory performance across all scenarios. To address this challenge, we construct three distinct classifiers, denoted as $f^{(1)}$, $f^{(2)}$, and $f^{(3)}$, each trained using a different optimization strategy.

\noindent \textbf{Reweight CE Loss.} The first classifier, $f^{(1)}$, is trained using a reweighted cross-entropy (CE) loss that explicitly compensates for class imbalance. The loss function is defined as:
\begin{equation}
    \mathcal{L}_{WCE} = - \sum_{y} w_y \cdot \mathbf{1}_{[y = \hat{y}]} \log p_y,
\end{equation}
where $y$ denotes the ground-truth label, $\hat{y}$ is the predicted label, $p_y$ is the predicted probability for class $y$. $w_y$ is the weight assigned to class $y$, calculated as inversely proportional to its frequency in the training data, \textit{i.e.}, $w_y = \frac{1}{f_y}$ with $f_y$ being the class frequency. This weighting mechanism compels the classifier to pay greater attention to the tail classes, which in our context often correspond to subtle or infrequent error patterns. As a result, $f^{(1)}$ is particularly effective in improving recall for underrepresented mistake instances.

\noindent \textbf{AUC Loss.} The second classifier, $f^{(2)}$, directly optimizes the Area Under the ROC Curve (AUC), which better reflects model performance in imbalanced classification settings. Specifically, the AUC loss is formulated as a pairwise ranking problem, aiming to assign higher prediction scores to positive (mistake) instances than to negative (correct) ones. The loss function is defined as:
\begin{equation}
    \mathcal{L}_{AUC} = \frac{1}{n^+n^-} \sum\limits_{i=1}^{n^+} \sum\limits_{j=1}^{n^-} \ell\left(f^{(2)}(\mathbf{F}_{\text{joint}}^{i,+}) - f^{(2)}(\mathbf{F}_{\text{joint}}^{i,-})\right),
\end{equation}
where $n^+$ and $n^-$ denote the numbers of mistake and correct samples respectively, $\mathbf{F}_{\text{joint}}^{i,+}$ and $\mathbf{F}_{\text{joint}}^{j,-}$ represent the joint features corresponding to the $i$-th mistake sample and the $j$-th correct sample. The function $\ell(\cdot)$ is typically a convex surrogate loss such as hinge or logistic loss. This formulation explicitly encourages the model to rank mistake segments above correct ones, thereby enhancing discriminative capability under long-tailed class distributions.

\noindent \textbf{LA Loss optimized with SAM.} The third classifier, $f^{(3)}$, leverages a combination of LA Loss and SAM to achieve improved generalization and robustness. Its objective is expressed as:
\begin{equation}
    \mathcal{L}_{LA} = \sum_{i=1}^{N} \ell\big((f^{(3)}(\mathbf{F}_{\text{joint}}^{i}) + \log \bf{p}), y_i \big), 
\end{equation}
where $\ell(\cdot)$ represents the cross-entropy loss, and $\bf{p} = [f_1, \cdots, f_C]$ denotes the class frequency vector. SAM further enhances this effect by seeking flat minima through an adversarial optimization scheme, formulated as:
\begin{equation}
    \min_{\theta} \max_{\Vert \epsilon \Vert \leq \rho} \mathcal{L}(f^{(3)}_{\theta+\epsilon})
\end{equation}
where the inner maximization identifies parameter perturbations within radius $\rho$ that locally worsen the loss. This forces the model to converge to parameter regions with improved generalization properties.

To unify the predictions from these heterogeneous classifiers, we introduce a Classification Mixture-of-Experts (C-MoE) module that assigns dynamic weights to the outputs of $f^{(1)}$, $f^{(2)}$, and $f^{(3)}$. The final prediction is computed as:
\begin{equation}
    \hat{y} = \sum_{k=1}^3 \beta_k f^{(k)}(\mathbf{F}_{\text{joint}}),
\end{equation}
where each $\beta_k$ is a learnable expert weight reflecting the relevance of the $k$-th classifier’s decision for the current input. This adaptive fusion enables the model to selectively emphasize the most suitable decision path for each instance, ultimately improving robustness and accuracy in complex mistake detection scenarios.
\section{Experiments}
\label{sec:experiments}
In this section, we describe some details of the experiments and present our results.

\subsection{Implementation Details}
\label{subsec:Implementation Details}
We conduct all experiments using eight NVIDIA A100 GPUs. For the feature extraction module, we adopt ViViT-B/$16\times 2$~\cite{arnab2021vivit} as our backbone and follow the original settings for hyperparameters. During fine-tuning, we apply LoRA~\cite{hulora} with a rank of $8$ to the attention layers, specifically targeting the query and value projection matrices. For the classification module, SAM~\cite{foretsharpness} is employed with a perturbation radius $\rho = 0.05$. The entire model is optimized using the Adam optimizer with a learning rate of $1 \times 10^{-5}$. We set the batch size to $128$ clips, each consisting of $32$ frames. The model is trained for a total of $10$ epochs.

\subsection{Results}
\label{subsec:Results}

\Cref{tab: results} presents the performance of various models on the mistake detection task. Compared to Random and TimeSformer, our method significantly improves the F-score. Furthermore, in comparison to the top-performing method of 2024, our method achieves a substantial improvement in mistake recall. Most notably, our method attains competitive performance using only the RGB modality, matching or even surpassing models that rely on multimodal inputs.

{
    \small
    \bibliographystyle{ieeenat_fullname}
    \bibliography{main}

\begin{thebibliography}{22}
\providecommand{\natexlab}[1]{#1}
\providecommand{\url}[1]{\texttt{#1}}
\expandafter\ifx\csname urlstyle\endcsname\relax
  \providecommand{\doi}[1]{doi: #1}\else
  \providecommand{\doi}{doi: \begingroup \urlstyle{rm}\Url}\fi

\bibitem[Arnab et~al.(2021)Arnab, Dehghani, Heigold, Sun, Lu{\v{c}}i{\'c}, and Schmid]{arnab2021vivit}
Anurag Arnab, Mostafa Dehghani, Georg Heigold, Chen Sun, Mario Lu{\v{c}}i{\'c}, and Cordelia Schmid.
\newblock Vivit: A video vision transformer.
\newblock In \emph{ICCV}, pages 6836--6846, 2021.

\bibitem[Bao et~al.(2024)Bao, Xu, Yang, He, Cao, and Huang]{bao2024improved}
Shilong Bao, Qianqian Xu, Zhiyong Yang, Yuan He, Xiaochun Cao, and Qingming Huang.
\newblock Improved diversity-promoting collaborative metric learning for recommendation.
\newblock \emph{PAMI}, 2024.

\bibitem[Cui et~al.(2019)Cui, Jia, Lin, Song, and Belongie]{cui2019class}
Yin Cui, Menglin Jia, Tsung-Yi Lin, Yang Song, and Serge Belongie.
\newblock Class-balanced loss based on effective number of samples.
\newblock In \emph{CVPR}, pages 9268--9277, 2019.

\bibitem[Foret et~al.(2021)Foret, Kleiner, Mobahi, and Neyshabur]{foretsharpness}
Pierre Foret, Ariel Kleiner, Hossein Mobahi, and Behnam Neyshabur.
\newblock Sharpness-aware minimization for efficiently improving generalization.
\newblock In \emph{ICLR}, 2021.

\bibitem[Han et~al.(2024)Han, Xu, Yang, Bao, Wen, Jiang, and Huang]{hanaucseg}
Boyu Han, Qianqian Xu, Zhiyong Yang, Shilong Bao, Peisong Wen, Yangbangyan Jiang, and Qingming Huang.
\newblock Aucseg: Auc-oriented pixel-level long-tail semantic segmentation.
\newblock In \emph{NeurIPS}, 2024.

\bibitem[Hanley and McNeil(1982)]{hanley1982meaning}
James~A Hanley and Barbara~J McNeil.
\newblock The meaning and use of the area under a receiver operating characteristic (roc) curve.
\newblock \emph{Radiology}, 143\penalty0 (1):\penalty0 29--36, 1982.

\bibitem[Hu et~al.(2022)Hu, Wallis, Allen-Zhu, Li, Wang, Wang, Chen, et~al.]{hulora}
Edward~J Hu, Phillip Wallis, Zeyuan Allen-Zhu, Yuanzhi Li, Shean Wang, Lu Wang, Weizhu Chen, et~al.
\newblock Lora: Low-rank adaptation of large language models.
\newblock In \emph{ICLR}, 2022.

\bibitem[Hua et~al.(2024)Hua, Xu, Bao, Yang, and Huang]{hua2024reconboost}
Cong Hua, Qianqian Xu, Shilong Bao, Zhiyong Yang, and Qingming Huang.
\newblock Reconboost: Boosting can achieve modality reconcilement.
\newblock In \emph{ICML}, pages 19573--19597, 2024.

\bibitem[Hua et~al.(2025)Hua, Xu, Yang, Wang, Bao, and Huang]{huaopenworldauc}
Cong Hua, Qianqian Xu, Zhiyong Yang, Zitai Wang, Shilong Bao, and Qingming Huang.
\newblock Openworldauc: Towards unified evaluation and optimization for open-world prompt tuning.
\newblock In \emph{ICML}, 2025.

\bibitem[Kay et~al.(2017)Kay, Carreira, Simonyan, Zhang, Hillier, Vijayanarasimhan, Viola, Green, Back, Natsev, et~al.]{kay2017kinetics}
Will Kay, Joao Carreira, Karen Simonyan, Brian Zhang, Chloe Hillier, Sudheendra Vijayanarasimhan, Fabio Viola, Tim Green, Trevor Back, Paul Natsev, et~al.
\newblock The kinetics human action video dataset.
\newblock \emph{arXiv preprint arXiv:1705.06950}, 2017.

\bibitem[Li et~al.(2024)Li, Xu, Bao, Yang, Cong, Cao, and Huang]{lisize}
Feiran Li, Qianqian Xu, Shilong Bao, Zhiyong Yang, Runmin Cong, Xiaochun Cao, and Qingming Huang.
\newblock Size-invariance matters: Rethinking metrics and losses for imbalanced multi-object salient object detection.
\newblock In \emph{ICML}, 2024.

\bibitem[Li et~al.(2025{\natexlab{a}})Li, Xu, Bao, Han, Yang, and Huang]{li2025hybrid}
Feiran Li, Qianqian Xu, Shilong Bao, Boyu Han, Zhiyong Yang, and Qingming Huang.
\newblock Hybrid generative fusion for efficient and privacy-preserving face recognition dataset generation.
\newblock \emph{arXiv preprint arXiv:2508.10672}, 2025{\natexlab{a}}.

\bibitem[Li et~al.(2025{\natexlab{b}})Li, Xu, Yang, Wang, Zhang, Cao, and Huang]{li2025focal}
Sicong Li, Qianqian Xu, Zhiyong Yang, Zitai Wang, Linchao Zhang, Xiaochun Cao, and Qingming Huang.
\newblock Focal-sam: Focal sharpness-aware minimization for long-tailed classification.
\newblock In \emph{ICML}, 2025{\natexlab{b}}.

\bibitem[Liang et~al.(2024)Liang, Zhu, Liu, Wu, Cao, and Chang]{liang2024badclip}
Siyuan Liang, Mingli Zhu, Aishan Liu, Baoyuan Wu, Xiaochun Cao, and Ee-Chien Chang.
\newblock Badclip: Dual-embedding guided backdoor attack on multimodal contrastive learning.
\newblock In \emph{CVPR}, pages 24645--24654, 2024.

\bibitem[Lin et~al.(2025)Lin, Li, Xu, Shi, Qin, Zhang, Sun, and Ji]{lin2025ltd}
Liuhao Lin, Ke Li, Zihan Xu, Yuchen Shi, Yulei Qin, Yan Zhang, Xing Sun, and Rongrong Ji.
\newblock Ltd-bench: Evaluating large language models by letting them draw.
\newblock \emph{NeurIPS}, 2025.

\bibitem[Liu et~al.(2024{\natexlab{a}})Liu, Jia, Xun, Liang, and Cao]{liu2024multimodal}
Xinwei Liu, Xiaojun Jia, Yuan Xun, Siyuan Liang, and Xiaochun Cao.
\newblock Multimodal unlearnable examples: Protecting data against multimodal contrastive learning.
\newblock In \emph{ACM MM}, pages 8024--8033, 2024{\natexlab{a}}.

\bibitem[Liu et~al.(2024{\natexlab{b}})Liu, Xu, Wen, Dai, and Huang]{liu2024not}
Yang Liu, Qianqian Xu, Peisong Wen, Siran Dai, and Qingming Huang.
\newblock Not all pairs are equal: Hierarchical learning for average-precision-oriented video retrieval.
\newblock In \emph{ACMMM}, pages 3828--3837, 2024{\natexlab{b}}.

\bibitem[Menon et~al.(2020)Menon, Jayasumana, Rawat, Jain, Veit, and Kumar]{menon2020long}
Aditya~Krishna Menon, Sadeep Jayasumana, Ankit~Singh Rawat, Himanshu Jain, Andreas Veit, and Sanjiv Kumar.
\newblock Long-tail learning via logit adjustment.
\newblock In \emph{ICLR}, 2020.

\bibitem[Wang et~al.(2023)Wang, Kwon, Rad, Pan, Chakraborty, Andrist, Bohus, Feniello, Tekin, Frujeri, et~al.]{wang2023holoassist}
Xin Wang, Taein Kwon, Mahdi Rad, Bowen Pan, Ishani Chakraborty, Sean Andrist, Dan Bohus, Ashley Feniello, Bugra Tekin, Felipe~Vieira Frujeri, et~al.
\newblock Holoassist: an egocentric human interaction dataset for interactive ai assistants in the real world.
\newblock In \emph{ICCV}, pages 20270--20281, 2023.

\bibitem[Yang et~al.(2021)Yang, Xu, Bao, Cao, and Huang]{yang2021learning}
Zhiyong Yang, Qianqian Xu, Shilong Bao, Xiaochun Cao, and Qingming Huang.
\newblock Learning with multiclass auc: Theory and algorithms.
\newblock \emph{PAMI}, 44\penalty0 (11):\penalty0 7747--7763, 2021.

\bibitem[Yang et~al.(2024)Yang, Xu, Wang, Li, Han, Bao, Cao, and Huang]{yang2024harnessing}
Zhiyong Yang, Qianqian Xu, Zitai Wang, Sicong Li, Boyu Han, Shilong Bao, Xiaochun Cao, and Qingming Huang.
\newblock Harnessing hierarchical label distribution variations in test agnostic long-tail recognition.
\newblock In \emph{ICML}, pages 56624--56664, 2024.

\bibitem[Zhang et~al.(2024)Zhang, Liu, Zhang, Liang, and Liu]{zhang2024towards}
Xinwei Zhang, Aishan Liu, Tianyuan Zhang, Siyuan Liang, and Xianglong Liu.
\newblock Towards robust physical-world backdoor attacks on lane detection.
\newblock In \emph{ACM MM}, pages 5131--5140, 2024.

\end{thebibliography}
}


\end{document}